\pdfoutput=1

\documentclass[11pt]{article}

\usepackage[final]{acl}

\usepackage{times}
\usepackage{latexsym}

\usepackage[T1]{fontenc}

\usepackage[utf8]{inputenc}

\usepackage{microtype}

\usepackage{inconsolata}

\usepackage[normalem]{ulem}
\newcommand{\draftcomment}[3]{}

\usepackage{amsfonts}
\usepackage{booktabs}
\usepackage{graphicx}
\usepackage{pifont}
\usepackage{amssymb}
\newcommand{\cmark}{\ding{51}}%
\newcommand{\xmark}{\ding{55}}
\newcommand{\toc}[0]{ToC}
\newcommand{\ctoc}[0]{Conceptual \toc}
\newcommand{\method}[0]{method}
\newcommand{\header}[0]{header}

\newcommand{\headers}[0]{headers}

\usepackage{mathtools}
\usepackage{bbm}
\usepackage{booktabs}
\usepackage{graphicx}
\mathchardef\mhyphen="2D

\usepackage{booktabs}
\usepackage{multirow}
\usepackage{graphicx}

\newcommand{\tenk}[0]{Form-10K}
\newcommand{\verds}[0]{verdict decisions}
\newcommand{\hebverds}[0]{Hebrew Verdict Decisions}

\title{Leveraging Collection-Wide Similarities for Unsupervised Document Structure Extraction}

\author{
Gili Lior$^{1,2}$~~~
Yoav Goldberg$^{1,3}$~~~
Gabriel Stanovsky$^{1,2}$
\\
$^1$Allen Institute for AI~~~
$^2$The Hebrew University of Jerusalem~~~
$^3$Bar-Ilan University
\\
\href{mailto:gili.lior@mail.huji.ac.il}{\texttt{gili.lior@mail.huji.ac.il}
}}
\begin{document}
\maketitle
\begin{abstract}


\emph{Document collections} of various domains, e.g., legal, medical, or financial, often share some underlying collection-wide structure, which captures information that can aid both human users and structure-aware models.
We propose to identify the typical structure of document within a collection, which requires to capture recurring topics across the collection, while abstracting over arbitrary header paraphrases, and ground each topic to respective document locations. 
These requirements pose several challenges: headers that mark recurring topics frequently differ in phrasing, certain section headers are unique to individual documents and do not reflect the typical structure, and the order of topics can vary between documents. 
Subsequently, we develop an unsupervised graph-based method which leverages both inter- and intra-document similarities, to extract the underlying collection-wide structure. 
Our evaluations on three diverse domains in both English and Hebrew indicate that our method extracts meaningful collection-wide structure, and we hope that future work will leverage our method for multi-document applications and structure-aware models.
\footnote{Our code and data are available at~\url{https://github.com/SLAB-NLP/Doc-Structure-Parser}}


\end{abstract}

\begin{figure*}[tb!]
    \centering
    \includegraphics[width=0.95\textwidth]{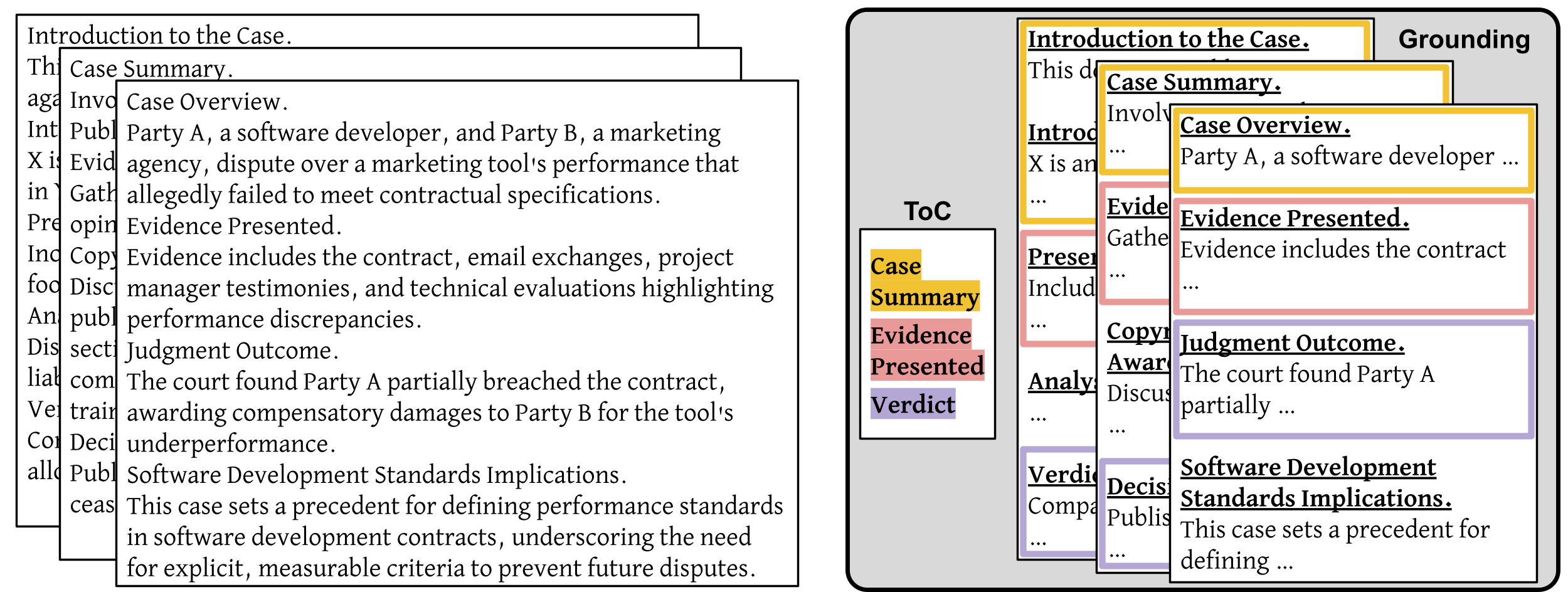}
    \caption{The input for our \method{} (on the left) a document collection with some shared underlying structure. The output (on the right) is a list of the most prominent topics across the collection (\toc), grounded to the documents, which is represented by the colored bounding boxes.}
    \label{fig:0}
\end{figure*}

\section{Introduction}

Knowing the structure of a typical document within a collection can be  useful in various use cases across different domains.
For example, consider the legal domain, where lawyers seek to analyze corpora of legal proceedings, looking for trends and patterns over time. In a retrieval scenario, they may look for punishment trends over many verdict decisions~\citep{wenger-etal-2021-automated}. While each verdict decision normally includes a dedicated punishment section, it is often hard to locate it, because it is not consistently marked -- different verdicts may use different \header{}s (``Punishment'', ``Sentencing Decision'', or ``Incurred Penalty''), and position the section in different document locations, requiring scholars to scan vast amounts of text. In an exploratory use case, lawyers may not have a well-defined apriori question, but instead interested in emerging collection-wide trends, for example, analyzing what judges take into account in verdict.


Furthermore, a collection-wide signal about document structure can also be leveraged by structure-aware models for multi-document downstream applications, e.g., infusing the document structure as part of the Transformer architecture, to improve multi-document applications~\cite{liu-lapata-2019-hierarchical,zhang-etal-2023-enhancing}. 


To support both human users and structure-aware models, we propose to identify the typical document structure within the collection (\S\ref{sec:task}). This requires to capture recurring topics across the collection, while abstracting over arbitrary \header{} paraphrases, and ground each topic to respective document locations. For example, in Figure~\ref{fig:0} we want to identify ``Case Summary'', ``Evidence Presented'' and ``Verdict'' as the recurring and prominent topics of a typical document within the collection of legal \verds{}, as opposed to ``Software Development Standards Implications'', which is document specific. This also requires to identify that ``Verdict'' and ``Judgement Decision'' represent the same topic, i.e., abstract over \header{} paraphrases. The colored bounding boxes in Figure~\ref{fig:0} represent the grounding unto each document.

Automatically extracting the typical document structure is challenging. While topic boundaries are often loosely defined by explicit \header{}s, it is hard to use them directly to understand collection-level properties, as headers indicating the same information often vary in phrasing, for example ``Verdict'', ``Judgement Outcome'' and ``Decision''. Moreover, some section headers are local to individual document, and do not participate in the global structure. For example, ``Software Development Standards Implication'' in Figure~\ref{fig:0}. Finally, while section order provides some signal, it is often inconsistent, and may vary across different documents. The challenge is then align section headers across the collection, while being flexible enough to discard sections that do not reflect a global shared structure.

Following, we devise an unsupervised method using a collection-wide signal to perform the structure extraction (\S\ref{sec:method}). To achieve this, we represent the document collection using a complete undirected weighted graph, whose nodes represent topic boundary candidates, and the weight of edges between each pair of nodes represents their semantic similarity. Such formulation allows us to model relations both within document, as well as across documents. 
For example, set large edge weight between ``Case Overview'' and ``Introduction to the Case'', as they convey semantically similar topics. 
Finally, we find communities within the graph, where each community is a group of similar nodes that form a single component of the collection-wide structure, and filter the communities that best outline the typical document structure, which we call the collection-wide table of contents (\toc{}).

To illustrate the robustness of our method across various domains and languages, we curate three distinct datasets (§\ref{sec:data}). These include two English datasets from different domains (financial and legal), and a Hebrew dataset, composed of legal documents. This variety shows that our method is adaptive and effective under different linguistic and domain-specific contexts.

To evaluate our model, we propose three distinct metrics (§\ref{sec:results}). First is the ``header intrusion''  human evaluation, which is adapted from the popular ``word intrusion'' metric for clustering evaluation~\citep{Chang2009ReadingTL}, to evaluate the quality of the representation of the collection. Second, an automatic evaluation for the document-level grounding, to evaluate the predicted grounding coverage. Last, a qualitative analysis over the predicted \toc{} entries, manually exploring how meaningful are the produced entries, comparing them to an existing \toc{} of the financial documents collection.

We find that our \method{} extracts meaningful typical document structure, capable of retrieving both the overall collection structure, while also mapping it unto individual documents, and that it is robust to varying domains and languages, with no supervision and little domain-specific adaptation. Detailed error analysis shows that while it is capable of mapping most topics unto documents, it struggles with identifying the exact topic boundaries. 

Our released code allows future work to leverage collection-wide structure to enrich both user-facing applications, and to integrate document structure within LLMs to cope with multiple documents.

Our main contributions are: (1) we formally define a novel task of identifying the typical document structure from a document collection, (2) we curate three datasets for the task from diverse domains and languages, and (3) we develop an unsupervised approach for the task, by leveraging collection-wide signal in a community detection algorithm.


\begin{figure*}[tb!]
    \centering
    \includegraphics[width=\textwidth]{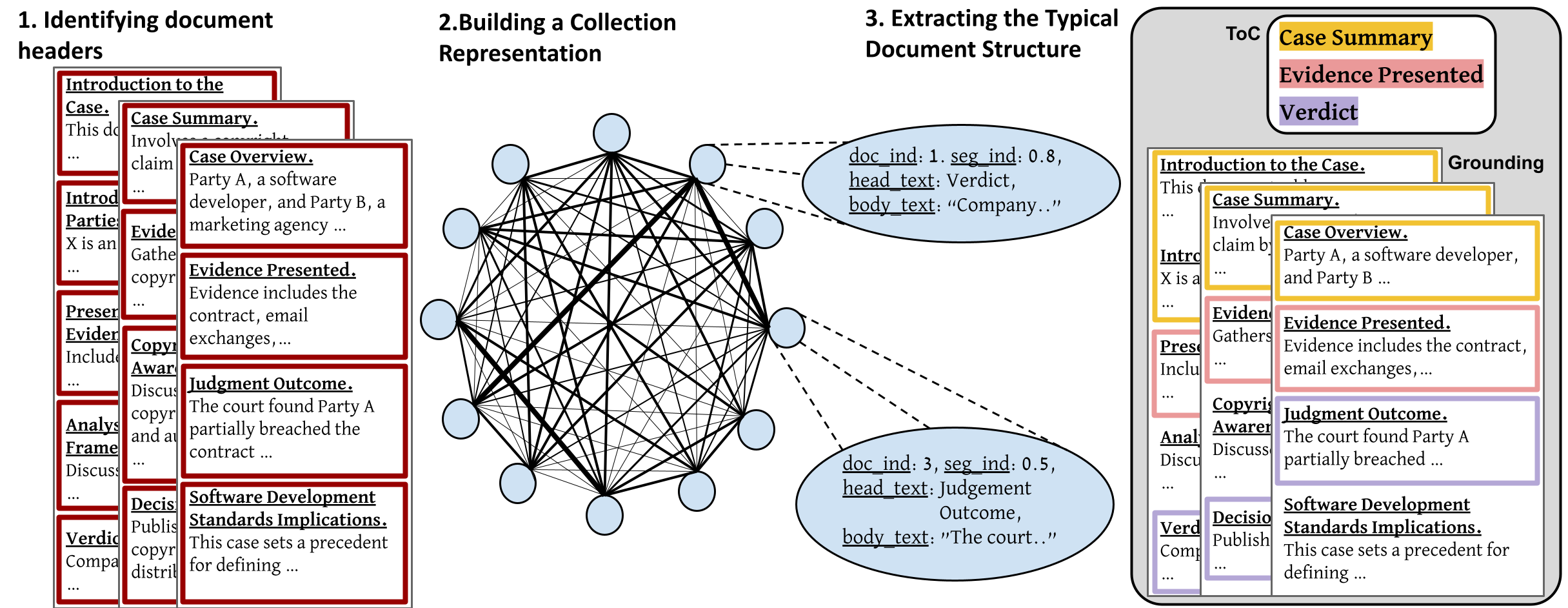}
    \caption{A high-level description of the three steps in our approach for extracting structure from a collection of documents. First, we identify topic boundaries in individual documents, based on lexical cues (\S\ref{sec:detect-headers}). Second, we build a collection representation using a complete graph whose nodes are the topics identified in the previous step, and weighted edges represent semantic similarity between them (\S\ref{sec:graph-construction}) Finally, we extract the typical document structure using unsupervised community detection, which we filter based on coverage maximization (\S\ref{sec:louvain}).}

    \label{fig:flowchart}
\end{figure*}

\section{Task Definition}\label{sec:task}


Given a collection of documents which share an underlying latent structure, we aim to recover a structure which satisfies the three following requirements. First, we would like to list \emph{the most dominant topics} which appear throughout the collection, and omit topics which appear only in few documents. This allows users to readily observe overall trends within the collection, without reading it in its entirety. Second, the representation should \emph{abstract over header paraphrases}, allowing users to identify that certain sections discuss similar semantic topics, despite being phrased differently. Third, the structure should \emph{ground topics unto the documents in the collection}, thus providing users with document-level structure, amenable for manual as well as automatic manipulation.

Formally, we denote a collection of documents $D=\{d_1, \ldots, d_n\}$, where a model is required to output a tuple $(T, M)$. $T = \{t_1, \ldots , t_k\}$ is a set of $k$ most prominent topics across $D$. In the scope of this work, each topic $t_i$ should appear as a continuous text span in some document $d_j$. 
For example, in Figure~\ref{fig:0}, $T=\{$``Case Summary'', ``Evidence Presented'', ``Verdict''$\}$.
$M:(T \times D) \mapsto (\mathbb{N}^{+}\times \mathbb{N}^{+})$ grounds topics to individual documents, mapping each topic to start and end indices. $M$ is exemplified with colored bounding boxes in Figure~\ref{fig:0}.

\section{Method}\label{sec:method}


We describe our \method{} for generating the \toc{} and grounding, which captures the typical document structure within a given document collection, and composed of three steps as outlined in Figure~\ref{fig:flowchart}. 

\subsection{Identifying Document Headers}\label{sec:detect-headers}

Following~\citet{erbs-etal-2013-hierarchy}, in the scope of this work we assume document-level topics appear as explicit headers within documents. Later we show that this assumption holds in real-world domains. 

Since header formatting may vary across domains, we consider their detection a corpus-specific task. We identify header candidates via rule-based heuristics, by leveraging a collection-wide signal that allows us to  differentiate between certain patterns which are prominent throughout the collection (e.g., sentence length and capitalization patterns~\citep{gutehrle2022processing}), while filtering out other document elements which are sometimes styled as headers (e.g., page numbers, recurring signatures, copyright licensing, etc.).







\subsection{Building a Collection Representation}\label{sec:graph-construction}

Next, we face a collection-level challenge. After segmenting each document to topics, it is not clear how to represent both intra- and inter-document similarities across the entire collection. Furthermore, the similarity between topics hinges on a myriad of factors, including semantic similarities between the topic headers, their content, and their location within the document. 
For example, in Figure~\ref{fig:flowchart} we want to identify that ``Verdict'' in one document and ``Judgement Desicion'' in another are both related to a shared `Verdict' topic, which is dominant in legal documents, and usually located towards the end of the verdict decision. 


To model the different intra- and inter-document similarities, we represent a document collection using a complete undirected weighted graph $G = (V, E)$ as elaborated below.

\paragraph{Graph nodes.} The nodes in the graph include all \headers{} identified previously. Each node $v \in V(G)$ is associated with a 4-tuple: \{\emph{doc\_ind, seg\_ind, head\_text, body\_text}\}. \emph{doc\_ind} $\in \mathbb{N}$ denotes the index of the document within the collection, \emph{seg\_ind} $\in [0, 1]$ denotes the \emph{normalized} sequential position within the document, while \emph{head\_text} and \emph{body\_text} are the two continuous text spans of the header and following body text. For example, one of the nodes in Figure~\ref{fig:flowchart} represents the segment at the $0.5$ normalized position from the 3rd document, its header is ``Judgement Outcome'' and its body text starts with ``The court..''.

\paragraph{Edge weights.}
As described above, $G$ is a complete graph over all identified topic boundary candidates from documents within the collection. For each edge $(v_1,v_2)$, we define $weight(v_{1}, v_{2}) \in \mathbb{R}$ as a weighted sum of three similarity metrics:
\begin{equation}
\label{eq:sim}
\begin{split}
     weight(v_{1}, v_{2}) = & \  \lambda_{head} \cdot head\_sim(v_{1}, v_{2}) \\
     &+ \lambda_{body} \cdot body\_sim(v_{1}, v_{2}) \\  
     & +  \lambda_{pos} \cdot pos\_sim(v_{1}, v_{2})
\end{split}
\end{equation}
Where $head\_sim(v_{1}, v_{2}) \in \mathbb{R}$ represents the similarity between the \headers{} corresponding to the two nodes,  and $body\_sim(v_{1}, v_{2}) \in \mathbb{R}$ represents the similarity between their bodies. We compute both using cosine similarities over a language model embedding. 
The $pos\_sim(v_{1}, v_{2}) \in \mathbb{R}$ similarity metric computes the similarity of ordering within the document, which encodes the assumption that documents sometimes follow similar topic ordering. In particular, we define $pos\_sim$ as follows:
\begin{equation}
\begin{split}
    pos\_sim&(v_{1}, v_{2}) = \\
    &\left(\left| seg\_ind(v_1) - seg\_ind(v_2) \right|\right)^{-1}
\end{split}
\end{equation}
Hence, $pos\_sim$ is larger the more the respective segments are in similar positions within their respective documents.

Finally, $\lambda_{head},\lambda_{body},\lambda_{pos}$ are hyperparameters, non-negative and sum up to $1$, weighting the three metrics according to different corpus characteristics.
For example, if the document collection follows a strict order across all documents, $\lambda_{pos}$ could be large, as opposed to the case that the order is not strict but the \headers{} themselves are similar, and then $\lambda_{head}$ would be larger.  

\begin{table*}[ht!]
\centering
\resizebox{\textwidth}{!}{%
\begin{tabular}{@{}lllllccl@{}}
\toprule
\textbf{Dataset Name} &
  \textbf{\# docs} &
  \textbf{\# words} &
  \textbf{Domain} &
  \textbf{Language} &
  \textbf{Strict structure} &
  \multicolumn{1}{l}{\textbf{Preserve sections order}} &
  \textbf{Headers similarity} \\ \midrule
\tenk          & 500 & 18M & Financial & English & \cmark & \cmark & High   \\
CUAD         & 389 & 3M & Legal (contracts)     & English & \xmark & \xmark & Medium \\
Heb-Verdicts & 277 & 3.4M & Legal (verdicts)    & Hebrew  & \xmark & \cmark & Medium \\ \bottomrule
\end{tabular}%
}
\caption{Description of the three datasets we use for evaluation, spanning different domains and languages, with varying properties, as detailed in Section~\ref{sec:data}.}
\label{tab:datasets}
\end{table*}

\subsection{Extracting the Typical Document Structure}\label{sec:louvain}
Finally, we want to find the the most representative collection-wide \toc{} and its corresponding grounding unto each document. 
At this step we want to find list of important topics which appear in many documents throughout the collection. 
In addition, we are interested in mapping the list of important topics unto each document in the collection.

As we outlined in Section~\ref{sec:task}, a good output would consist of coherent topics (e.g., all topics mapped to "Case Summary" would discuss the facts of the case), covering as many documents in the collections, and much of each individual document.


We achieve this by finding communities in the graph. Each community is a group of nodes heavily connected to each other by high-weight edges and lightly connected to nodes outside the community by low-weight edges. Ideally, the inter-community edges represent high similarity between nodes, and hence the community represents a coherent topic across different documents within the collection.




For community detection, we apply the Louvain algorithm~\cite{blondel2008fast}. We use this algorithm as it has a small number of hyper-parameters compared to other community detection algorithms, and it has shown reliable results over various NLP tasks~\cite{lucy-etal-2023-words}. The objective of the Louvain algorithm is to maximize modularity, which correlates with the ratio between edges inside the community to edges outside the community.

Finally, we generate the collection-wide \toc{} out of the best communities. We find a subset of $k$ communities that maximize the coverage of the entire collection, which indicates that the chosen communities represent the topics that are relevant to many of the documents and covers the most of each document. For example, in Figure~\ref{fig:0} we omit ``Software Development Standards Implications'' as it appears only in a single document.

Formally, let $C$ represent the set of all communities output by Louvain, where each community $c\in C$ is a list of nodes. We want to find the subset $S\subset C$, containing $k$ communities, that maximizes the coverage  of each individual document $d$ across the entire collection $D$:
\begin{equation}
    \underset{S\subset C \; s.t.\; |S|=k}{\mathrm{argmax}} \:\: 
    \sum_{d\in D}\left[\frac{1}{\left\vert{d}\right\vert} \sum_{v\in d} \mathbbm{1}_{[\exists c \in S\;s.t.\; v\in c]}\right]
\end{equation}


To form the $k$ \toc{} entries, we then find the centroid of each of the $k$ best communities, and ground each representative centroid to the text spans of its corresponding community. 


\section{Data Curation}\label{sec:data}

To evaluate our \method{}, we curate three document collections. As presented in Table~\ref{tab:datasets}, these collections cover different domains and languages, with varying structure properties, showing the robustness of our \method{}.

\subsection{\tenk}
\tenk{} is an annual financial report required by U.S. Securities and Exchange Commission, summarizing a company's annual financial performance. It includes information such as company history, organizational structure, executive compensation, equity, and audited financial statements.

The \tenk{} files were collected from the SEC EDGAR platform.\footnote{Using the secedgar python package~\url{https://github.com/sec-edgar/sec-edgar}} We sampled 500 company CIK tickers from a list of public companies, and used these as the basis for EDGAR form retrieval. We then extracted the document's texts by cleaning the markup and filtering non-textual lines (e.g. tables) using regular expressions.

\subsection{CUAD}

~\citet{hendrycks2021cuad} presented Contract Understanding Atticus Dataset (CUAD), consisting of 510 commercial legal contracts.
This dataset was curated to evaluate the extraction of key clauses in contracts, providing labeled clauses that were manually extracted from the documents. Since we are only interested in predicting the structure of the documents, we ignore the provided labels. 

To run our \method{} over this dataset, we converted the provided pdf files to raw texts, resulting in raw texts of 389 legal contracts.\footnote{We used Adobe API package to convert pdf to text \url{https://developer.adobe.com/document-services/docs/overview/pdf-extract-api/}} 

In contrast with \tenk{}, CUAD consists of different types of contracts, such as affiliate agreements, license agreements, marketing, or manufacturing. Hence, the contracts may diverge in their structure, content and order. 

\subsection{\hebverds{}}
We use ~\citet{habba2023perfect}'s curated dataset of \verds{} from the Israeli court, in Hebrew. The original dataset consists of 855 \verds{} in sexual offense cases, which we then filtered manually, keeping only the documents with explicit headers, resulting in a collection of 277 Hebrew \verds{}.

Since all \verds{} in this dataset are of the same domain (sexual assault cases), we expect them to contain some overlapping topics and high-level structure similarities. But, since there is no pre-defined structure and each judge has their personal style, there are also expected differences and some parts may be added or omitted. 






\section{Results}\label{sec:results}

We evaluate three different aspects of our \method{}, including human evaluation of the communities we produce (§\ref{sec:intruder-eval}), document-level evaluation of the \toc{} grounding (§\ref{sec:doc-level-eval}), and a qualitative analysis of the predicted \toc{} (§\ref{sec:10k-qualitative}). 

\subsection{Experimental Setup}\label{sec:setup}


Our architecture introduces several hyperparmeters designed to capture different datasets properties.  
Below we outline their configuration for each of our three collections. These are summarized in Table~\ref{tab:datasets}, and presented in detail in Table~\ref{tab:params} in the Appendix.

First, our \method{} embeds the document texts using a pre-trained language model. For the two English datasets we use a version of the MPNet language model~\cite{NEURIPS2020_c3a690be}, as it was reported with the highest average performance on sentence embedding and semantic search tasks.\footnote{The MPNet version we use ~\url{https://huggingface.co/sentence-transformers/all-mpnet-base-v2}}$^,$\footnote{Performance of different pre-trained English LMs~\url{https://www.sbert.net/docs/pretrained_models.html}}
For our \hebverds{} dataset we use a version of AlephBERT~\citep{seker2021alephbert}.\footnote{The AlephBERT version we use~\url{https://huggingface.co/imvladikon/sentence-transformers-alephbert}}

Second, we configure the weighting of the different similarity metrics in Equation~\ref{eq:sim},  according to apriori knowledge of the domains.
For example, we set a high header similarity weight ($\lambda_{head}$) for the \tenk{} corpus, to reflect the expectation for very similar headers across documents. In contrast, the structure of the documents in CUAD is more flexible, so we set a more uniform weighting.

\subsection{Graph Representation Evaluation}\label{sec:intruder-eval}
 
In this evaluation, participants are shown 10 headers, 9 of which are chosen headers from the same community at random, and the last header is chosen at random from outside the community.  Participants are then tasked with identifying the intruder.

This task assesses if our predicted communities are well separated. We expect meaningful communities to demonstrate high internal similarity and low similarity with other communities. Therefore, if our communities are well-defined and meaningful, identifying intrusions should be straightforward, as they will noticeably differ from the rest of the community. 

Our evaluation is an adapted version of the ``word intrusion'' task~\citep{Chang2009ReadingTL},
widely used for human evaluation of topic clustering~\citep{Bhatia2017AnAA, Bhatia2018TopicIF, Prouteau2022AreES}.
We adapt this metric to account for the higher variability presents in headers as opposed to single-word topics by allowing annotators to choose an arbitrary number of intruders. The number of options they mark is reflected in a \emph{confidence} metric:
\begin{equation}\label{eq:conf}
    conf=1-\frac{num\_marked-1}{num\_options}
\end{equation}
Intuitively, the more intruders that the annotator marks, the less they are certain of their annotation. At the extremes, $conf = 1$ when an annotator marks a single intruder, and $conf = \frac{1}{10}$ if they mark all options as intruders. This metric does not account for the accuracy of the annotation, which is calculated separately, if either any of the marked options is the  actual intruder.

\begin{table}[t]
\centering
\resizebox{\columnwidth}{!}{%
\begin{tabular}{@{}lll@{}}
\toprule
 & Accuracy & \multicolumn{1}{c}{Confidence} \\ \midrule
\tenk & $67.5$ & $85.6 \pm 22.4$ \\
CUAD & $61.7$ & $77.8 \pm 21.3$ \\
\hebverds & $65.2$ & $89.2 \pm 21.9$ \\ \midrule
\begin{tabular}[c]{@{}l@{}}Randomly choosing \\ 3 candidates\end{tabular} & $30.0$ & $80.0$ \\ \bottomrule
\end{tabular}%
}
\caption{Header intrusion crowdsource human evaluation. Confidence is determined based on how many options the participant marked as an intruder candidate according to Equation~\ref{eq:conf}, presenting the average $\pm$ std confidence across all annotations. Accuracy is gained if either one of the marked candidates is the actual intruder. The last row is a random baseline of choosing three random intruder candidates, which induces $80\%$ confidence, resulting in expected accuracy of $30\%$. 
}
\label{tab:header-intrusion}
\end{table}

\paragraph{Crowdsourcing configuration.}

We run this evaluation through Amazon Mechanical Turk, with a pool of $12$ participants, where each unique sample was annotated by exactly one participant, for a total of $900$ collected annotations (around $450$ samples for each English dataset). We pay $0.1\mhyphen0.2$ USD  for a single annotation, aiming for an hourly pay of $12$ USD.
To ensure quality annotations, we start with a qualification test where participants need to reach accuracy above $0.4$ with confidence above $0.25$ over a few examples of our method's output. 
In Figure~\ref{fig:intruder-annotation} in the Appendix we provide an example of the annotation interface. As for the \hebverds, we collect $132$ annotations via a group of $8$ in-house Hebrew-speaking graduate students, who show similar performance to MTurk qualified workers.

\subsubsection{Key Findings}

In Table~\ref{tab:header-intrusion} we present the results for the \textit{header intrusion} human evaluation, and highlight key findings below.

\paragraph{Our \method{} predicts a meaningful collection representation on par with traditional topic modelling.}
In Table~\ref{tab:header-intrusion} we see that the accuracy for intrusion detection is more than double the chance accuracy for the observed $\sim$$80\%$ confidence. Moreover,  we find that our reported accuracy falls inside ~\citet{Chang2009ReadingTL}'s interquartile range (IQR) of word-intrusion results.

\paragraph{Our \method{} performs better on document collections with a more strict nature of structure.} 
As seen in Table~\ref{tab:header-intrusion}, accuracy is highest for the \tenk{} collection, which may attributed to its stricter structure, compared to our other two collections.


\begin{table*}[t]
\centering
\resizebox{0.8\textwidth}{!}{%
\begin{tabular}{@{}l|llll|llll@{}}
\toprule
 & \multicolumn{4}{c|}{\textbf{\tenk}} & \multicolumn{4}{c}{\textbf{\hebverds}} \\
 & \multicolumn{2}{c}{Macro F1} & \multicolumn{2}{c|}{Micro F1} & \multicolumn{2}{c}{Macro F1} & \multicolumn{2}{c}{Micro F1} \\
 & partial & exact & partial & exact & partial & exact & partial & exact \\ \midrule
Our \method & 93.7 & 63.7 & 92.6 & 43.9 & 78.1 & 63.7 & 82.8 & 64.2 \\
Most frequent class & 6.3 & 0 & 21.5 & 0 & 8.3 & 0 & 18.7 & 0 \\
Random & 16.1 & 1.1 & 25.8 & 1.1 & 21.6 & 6.4 & 16.6 & 3.2 \\ \bottomrule
\end{tabular}%
}
\caption{The accumulated \toc{} grounding scores for the \tenk{} and \hebverds{} datasets, comparing our \method{} to two different baselines, as elaborated in §\ref{sec:doc-level-eval}.}
\label{tab:f1-baseline}
\end{table*}

\paragraph{Our \method{} sometimes conflates opposite headers from opposite topics.} We manually analyze the \hebverds{} annotations, finding that some of the low confidence annotations (i.e., when annotators mark many options for the intruder) happen when our \method{} conflates opposite topics in the same community. For example, ``defense arguments'' and ``prosecution arguments'' are wrongly clustered in the same community.\footnote{All examples from \hebverds{} are translated from Hebrew.} Considering opposite topics as close in the embedding space is a known property of language models~\cite{vahtola-etal-2022-easy}.

\subsection{Document Grounding Evaluation}\label{sec:doc-level-eval}
In a second evaluation effort, we aim to quantify the quality of our \method{}'s automatic grounding from collection-wide \toc{} topics unto text spans in individual documents.

\paragraph{Annotating gold labels.} 
To evaluate our \method{}'s grounding performance, we annotate some of our unsupervised data in two of our domains. The \tenk{} dataset follows a strict structure, so we apply simple heuristics to automatically generate gold label grounding, resulting in a collection of 266 documents with grounding gold labels.\footnote{Gold \toc{} for \tenk{} is  based on \url{https://en.wikipedia.org/wiki/Form_10-K}} In contrast, for the \hebverds{} we use human annotations to generate its grounding gold labels, done by a domain expert (Hebrew speaking law student). Each manual annotation took approximately $45$ minutes, with a total of $15$ labeled documents. A screenshot of the annotation interface is provided in Figure~\ref{fig:doc_grounding} in the Appendix. We omitted CUAD from this particular evaluation due to budget constraints.

\paragraph{Metrics.}
We formulate exact and partial match scores, evaluating the overlap between the gold and the predicted grounding. Partial match credits \emph{any} overlap between the text spans in predicted and gold, while exact match counts only \emph{perfectly} aligned predictions.  The formal mathematical definitions are presented in Appendix~\ref{sec:appendix-match}.


\paragraph{Baselines.}
To the best of our knowledge, we present the first \method{} which aims to identify \toc{} at the document-collection level, and hence we are not aware of applicable baselines. Instead, to gauge the difficulty of this task we include two naive approaches. First, a baseline predicting \textit{most frequent class} in the gold annotations. For the \tenk{} corpus this is ``Exhibits, Financial Statement Schedules'' ($22\%$ of headers) and for \hebverds{} this is ``prosecution evidence'' ($20\%$ of headers). The second baseline predicts each gold \toc{} header uniformly \emph{at random}, averaged over 100 different random seeds.

\subsubsection{Key Findings}
The results for our two metrics and two document collections are presented in Table~\ref{tab:f1-baseline}. Several observations can be drawn based on these results.

\paragraph{Our \method{} captures the correct structure of a document.} The high partial match scores indicate that our \method{} is able to predict the correct document structure, as  almost every gold topic grounding has some overlap with the predicted topic grounding (Micro F1 score of $92.6$). As shown in Table~\ref{tab:f1-baseline}, our \method{} outperforms the two naive baselines in all F1 scores.

\paragraph{Exact match score is noisy.} Even if only a single sentence is separating between a gold and a predicted topic grounding boundaries, the exact match score for the topic will be $0$. We find this score very strict, and somewhat arbitrary. In addition, topics related to longer text spans, are more probable to reach a low exact match score. This is indicated by the first three rows in Table~\ref{tab:precision-recall} in the Appendix, representing the three largest classes, showing lower exact match scores than all other classes. 

\begin{figure*}[tb!]
    \centering
    \includegraphics[width=0.95\textwidth]{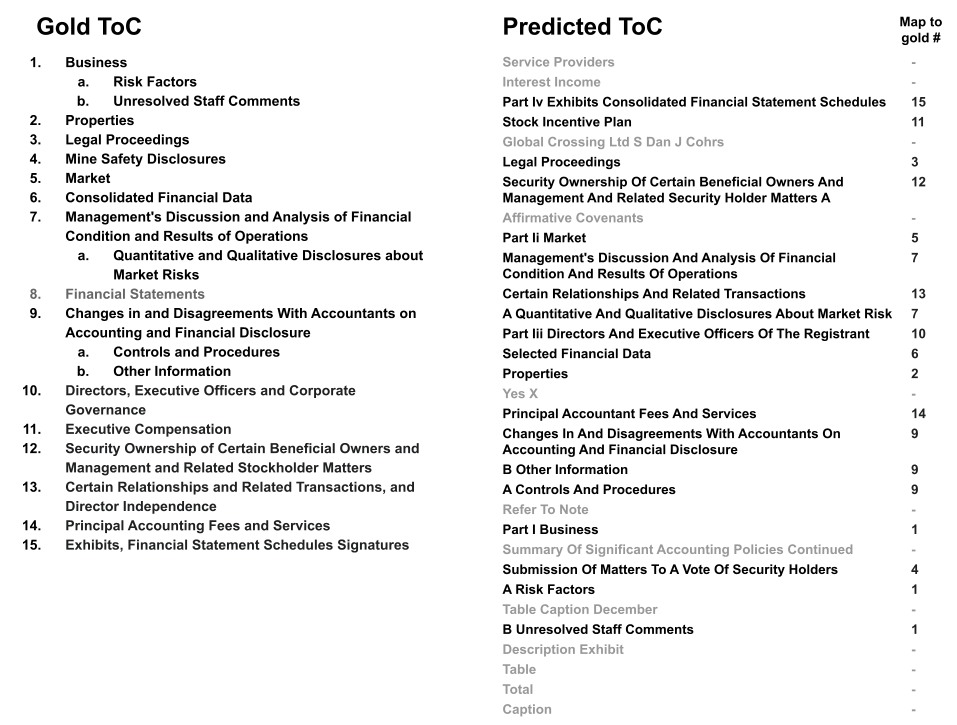}
    \caption{Mapping between the gold and predicted \toc{} for the \tenk{} dataset. The predicted \toc{} entries are ordered according to communities coverage ranking (§~\ref{sec:louvain}). The entries in gray does not have mapping.}
    \label{fig:toc}
\end{figure*}

\subsection{Qualitative \toc{}  Analysis}\label{sec:10k-qualitative}
Lastly, we suggest a collection-level qualitative evaluation, to explore the quality of the output \toc. Below we provide several key observations derived from our manual mapping between the predicted and gold \tenk{} \toc{} entries, visualized in Figure~\ref{fig:toc}. 

We were able to map 14 out of the 15 gold \toc{} entries to the predicted \toc{} entries. The only gold entry that was not mapped is ``8. financial statements''. After manually exploring the \toc{} grounding, we find that our \method{} sometimes confuses this topic with ``6. Consolidated Financial Data''. 

The 14 gold \toc{} entries align with 19 predicted entries, in the following manner:
\begin{itemize}
    \itemsep0em 
    \item 13 predicted \toc{} entries that were textually almost identical to the gold entries, with a straight-forward mapping.
    \item 1 mapping which followed a manual exploration of the dataset -- we mapped ``4. Mine Safety Disclouser'' to ``Submission Of Matters To A Vote Of Security Holders'', as we find that in many of the documents the latter is the actual \header{} of the $4^{th}$ topic.
    \item 5 \emph{subsection} headers (`A Risk Factors', `B Unresolved Staff Comments', etc.), that we map to their gold topic \headers, as our \method{} predicts a flat structure while the \tenk{} consists of an hierarchical structure.
\end{itemize}

Beyond the 19 predicted entries that align with gold entries, our \method{} predicted additional 12 entries.
Some topics indeed recur throughout the collection and accordingly receive high coverage scores, e.g., ``Service Providers'' or ``Interest Income'', and may be considered too granular. Others, attributed to noise, e.g., ``Total'', ``Table'' or ``Yes X'', receive low coverage scores, and their emergence as topic candidates could stem from degraded performance at the header detection step.

\section{Related Work}\label{sec:related_work}





A seminal theory of a pragmatic meaning is rhetorical structure theory~(RST~\citealp{mann1988rhetorical}), that inspired much follow-up work,  mostly approaching document structure extraction as a supervised task~\cite{arnold-etal-2019-sector,kang-etal-2022-financial,10.1145/3477495.3531817, hua-wang-2022-efficient}. The need for supervision narrows the application of such tools only to high-resource languages and small number of domains. Our method is unsupervised, does not require labeling, and is applicable to many languages and domains.

Another line of work suggest to approach structure extraction as an unsupervised task, to extend its applicability~\cite{xing-carenini-2021-improving, born-etal-2022-sequence, li-etal-2023-discourse}, but they still approach this as a single-document task. In our work we take the entire collection as our input, leveraging signal regarding the underlying collection-wide structure.

\citet{xing-etal-2022-improving} discovered that injecting external (above-sentence) information can help with identifying discourse dependency structure, but they only look at individual documents. We extend this observation into modeling relations across an entire collection, to inject external information about the structure components of the collection. 

Finally, theories like CST~\cite{radev-2000-common} and CAST~\cite{9099835}, or~\citep{Lu2018FunctionalSI}, suggest to look at the entire data collection to extract the main components that capture a typical structure layout of a document within the collection, but, their approach too relies on supervision and labeled data.

\section{Conclusion}

In this paper we suggested an unsupervised \method{} to extract the typical document structure within a collection, leveraging similarities across documents that come from the same collection with some shared underlying latent structure.


We evaluated the extracted typical document structure through three different evaluation metrics, over three different datasets, showing our method's robustness to different domains and languages. The evaluations indicated that our method succeeds in identifying the different topics consisting the typical document structure, while showing room for improvement in its accuracy to detect the exact location of where a topic starts and where it ends.

Our method can be leveraged by users of multi-document applications, allowing them more focused browsing over a collection of documents, which can be useful for downstream tasks like retrieval and summarization. As for models, future work can utilize our method and inject the collection-wide structure to structure-aware models, to improve downstream applications like multi-document summarization or cross document information extraction.

\section*{Acknowledgements}
We would like to thank the anonymous reviewers for their helpful comments. Members of the SLAB group at the Hebrew university for their productive discussion. 
This research was supported by grants from Allen Institute for AI,  the Israeli Ministry of Science and Technology (grant no. 2088), and the Council for Higher Education.

\section*{Limitations}

Currently, our method relies on segmentation induced by the \headers{} in the document, which limits our work only applicable on documents with explicit \headers. 
Future work can extend our collection-wide approach to develop a multi-document unsupervised method to induce segmentation for document collections without explicit \headers{}.

Furthermore, our method only predicts a flat list of the prominent topics, and does not express the hierarchical nature of structure that might be present. Future work can extend our method to predict also hierarchical structure.

\section*{Ethics Statement}

All annotators were informed ahead that their annotations are curated for research purposes, and how the data is going to be used.

The \hebverds{} dataset contains sensitive information pertaining to sexual harassment offenses. The original dataset we utilized underwent a rigorous anonymization process to protect individual privacy. We do not publicly publish this dataset, and accessing it requires authorization from the dataset's original proprietors~\cite{habba2023perfect}. 

Recognizing the potential for distress by annotating the \hebverds{}, we issued a trigger warning for all annotation tasks and obtained explicit consent from annotators before their participation. This ensured they were fully informed of the risks associated with exposure to sensitive and potentially triggering content.

\bibliography{custom}

\appendix

\section{Parameters For Evaluation}
\label{sec:appendix-params}

As mentioned in Section~\ref{sec:setup}, below we describe the different parameters used for running the evalution.

\begin{table}[ht!]
\centering
\resizebox{\columnwidth}{!}{%
\begin{tabular}{@{}lllll@{}}
\toprule
Dataset &
  Language Model &
  $\lambda_{head}$ &
  $\lambda_{body}$ &
  $\lambda_{index}$ \\ \midrule
10k  & all-mpnet-base-v2 & 0.7 & 0   & 0.3 \\
CUAD & all-mpnet-base-v2 & 0.5 & 0.3 & 0.2 \\
Heb-Verdicts &
  \begin{tabular}[c]{@{}l@{}}sentence-transformers-\\ alephbert\end{tabular} &
  0.5 &
  0.25 &
  0.25 \\ \bottomrule
\end{tabular}%
}
\caption{The implementation parameters we used for running our method on different data collections.}
\label{tab:params}
\end{table}

\paragraph{Headers detection.} In our implementation we detect headers for each dataset with a different set of heuristics. For example, for the Form-10k dataset, we search for sentences shorter than 10 tokens, with capitalization over > 50\% of the words in the sentence. This produces high recall (most headers are caught through this heuristic), but low precision (many of the caught header candidates are not really headers). An example for a scenario where a sentence is wrongly marked as a header candidate is for missing sections, in which the only section content is ‘None’ or ‘Omitted’. Furthermore, some of the header candidates are real headers, but not informative ones, for example ‘Section A’. In such cases, we also drop these sentences from the set of header candidates using regex.

\section{Additional Evaluation Analysis}

\subsection{Exact and Partial Match Scores}\label{sec:appendix-match}

Topic boundaries are a set of consecutive text spans that are labeled with the same \toc{} header. Following, we define an \textit{exact match} of a topic grounding if the text spans forming the gold label topic grounding, match perfectly with the text spans that are predicted with that header, while for \textit{partial match} it is enough if there is one text span within that topic boundaries that was predicted correctly.

These two matching scores are formalized as follows: let $s(w,d)$ be the range of text spans in document $d$, that are labeled as class $w$. So, we evaluate the intersection between $s_{gold}(w,d)$ and $s_{pred}(w,d)$. Full match precision is if $s_{pred}(w,d)\setminus s_{gold}(w,d)= \emptyset$, i.e., all text spans in $d$ that were predicted $w$ are labeled correctly, and full match recall is if $s_{gold}(w,d)\setminus s_{pred}= \emptyset$, i.e., all text spans in $d$ that are labeled $w$ are covered by the correct predictions. For partial match it is enough to have a non-empty intersection between $s_{gold}(w,d)$ and $s_{pred}(w,d)$.

\subsection{Document-Level Grounding For \tenk}

\begin{table*}[t]
\resizebox{\textwidth}{!}{%
\begin{tabular}{@{}lllllllll@{}}
\toprule
\multirow{2}{*}{Gold Header} &
  \multicolumn{2}{c}{Class Size} &
  \multicolumn{2}{c}{Precision} &
  \multicolumn{2}{c}{Recall} &
  \multicolumn{2}{c}{F1 Score} \\
                                        & \# segments & \# sections & partial & exact & partial & exact & partial & exact \\ \midrule
Exhibits, Financial Statement Schedules & 3386        & 210         & 77.4    & 2.6   & 97.6    & 17.1  & 86.3    & 4.6   \\[0.1cm] 
\begin{tabular}[c]{@{}l@{}}Management s Discussion and Analysis of \\ Financial Condition and Results of Operations\end{tabular} &
  2556 &
  254 &
  95.1 &
  38.5 &
  99.2 &
  27.6 &
  97.1 &
  32.1 \\[0.35cm] 
Business                                & 1833        & 207         & 83.3    & 65.7  & 93.7    & 17.4  & 88.2    & 27.5  \\[0.1cm] 
\begin{tabular}[c]{@{}l@{}}Certain Relationships and Related Transactions, and \\ Director Independence\end{tabular} &
  1509 &
  257 &
  96.9 &
  75.2 &
  97.3 &
  82.9 &
  97.1 &
  78.8 \\[0.35cm] 
Principal Accounting Fees and Services  & 1039        & 210         & 93.6    & 83.2  & 98.1    & 71.9  & 95.8    & 77.1  \\[0.1cm] 
\begin{tabular}[c]{@{}l@{}}Changes in and Disagreements With Accountants on \\ Accounting and Financial Disclosure\end{tabular} &
  665 &
  238 &
  90.5 &
  64.6 &
  100 &
  79.8 &
  95.0 &
  71.4 \\[0.35cm] 
Executive Compensation                  & 452         & 255         & 97.3    & 35.2  & 99.6    & 87.1  & 98.4    & 50.2  \\[0.1cm] 
Market                                  & 400         & 247         & 96.3    & 73.6  & 94.3    & 71.3  & 95.3    & 72.4  \\[0.1cm] 
Legal Proceedings                       & 398         & 260         & 98.9    & 83.7  & 100     & 88.1  & 99.4    & 85.8  \\[0.1cm] 
Properties                              & 362         & 253         & 98.7    & 94.8  & 89.7    & 78.7  & 94.0    & 86.0  \\[0.1cm] 
\begin{tabular}[c]{@{}l@{}}Security Ownership of Certain Beneficial Owners and \\ Management and Related Stockholder Matters\end{tabular} &
  335 &
  247 &
  95.0 &
  73.1 &
  100 &
  88.3 &
  97.4 &
  80.0 \\[0.35cm] 
Directors, Executive Officers and Corporate Governance &
  327 &
  218 &
  88.6 &
  69.9 &
  95.9 &
  76.1 &
  92.1 &
  72.9 \\[0.1cm] 
Selected Financial Data                 & 320         & 241         & 97.9    & 82.6  & 97.9    & 87.1  & 97.9    & 84.8  \\[0.1cm] 
Mine Safety Disclosures                 & 292         & 220         & 93.5    & 89.0  & 65.9    & 55.5  & 77.3    & 68.3  \\ \bottomrule
\end{tabular}%
}
\caption{\toc{} grounding scores for the Form 10k dataset. Each row represents a single \toc{} entry, with its size within the dataset, and our \method's precision, recall and F1 scores on each class. The partial/exact match scores differ in their minimal requirement to classify a prediction as correct. While partial score is gained if the prediction and label has at least one overlapping segment, exact match is gained only if the prediction and gold label fully agree. We further elaborate on partial/exact match scores in Section~\ref{sec:doc-level-eval}. }
\label{tab:precision-recall}
\end{table*}

In Table~\ref{tab:precision-recall} we provide our \method 's detailed results for each \toc{} entry's grounding, i.e., precision, recall and F1 exact and partial scores, over the \tenk{} dataset. 

\subsection{Mapping Between Predicted and Gold \tenk{} \ctoc}

In Figure~\ref{fig:toc} we provide the manually mapping between the predicted \tenk{} \toc{} entries, to its gold label \toc.

\section{Human Annotations}
\label{sec:appendix-intrusion}

\subsection{Annotating Gold Labels For Grounding}

\begin{figure*}[t]
    \centering
    \includegraphics[width=\textwidth]{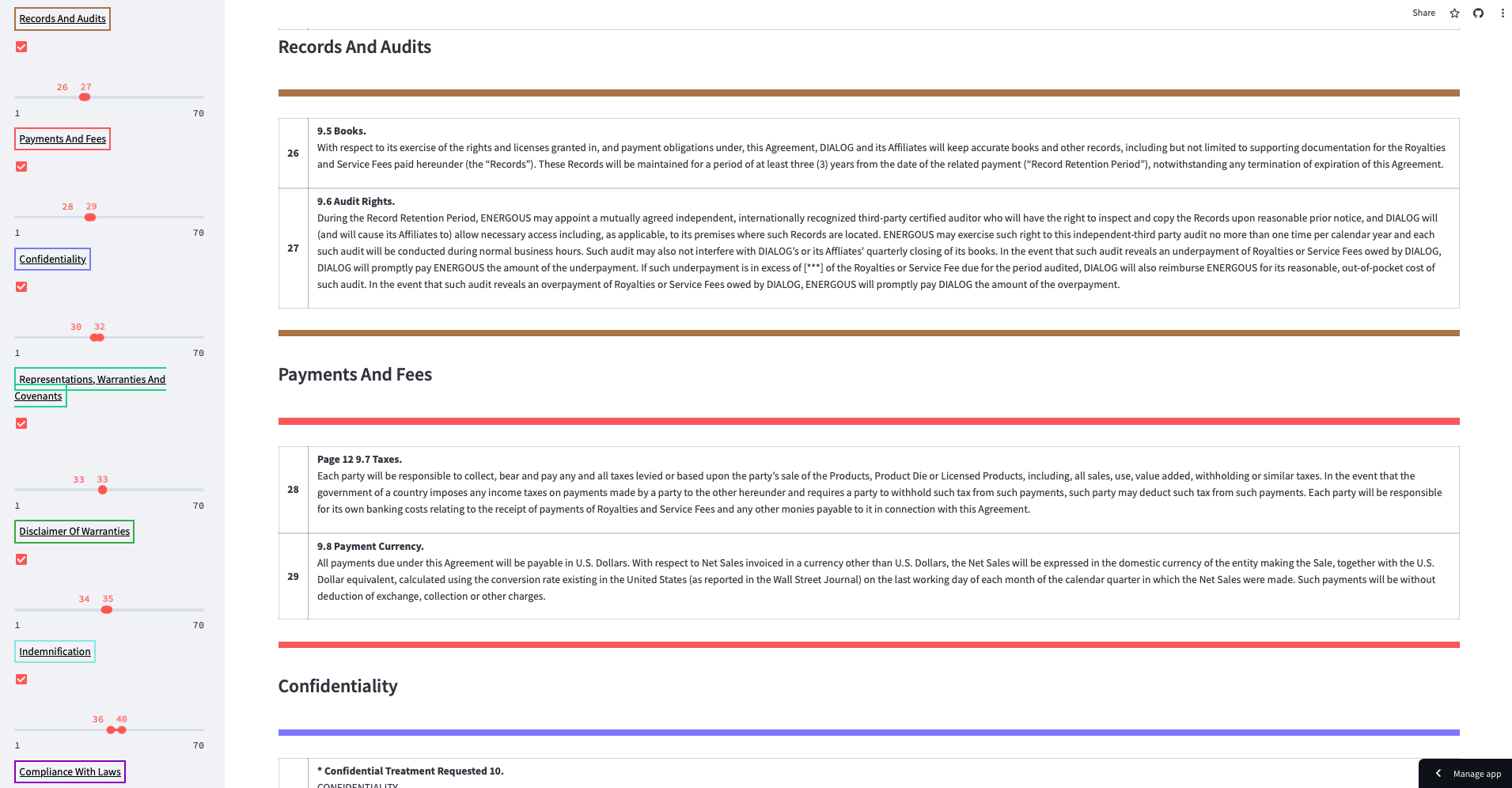}
    \caption{Example of the annotation interface for document-level grounding. On the left side is the \toc{} entries, and the bounding boxes on the around the text content marks the grounding boundaries. Below the \toc{} entries are editable bars, which than the annotator can edit according to its belief gold labeling.
    The presented document is partial, due to visbility limitation, but annotator can scroll and see the entire document.}
    \label{fig:doc_grounding}
\end{figure*}

Figure~\ref{fig:doc_grounding} is a screenshot of the annotation interface for grounding gold labels. We use this interface to annotate the \hebverds{} dataset, but for readability we provide here the interface over a document from CUAD.

\subsection{Header Intrusion Task}
\begin{figure*}[t]
    \centering
    \includegraphics[width=\textwidth]{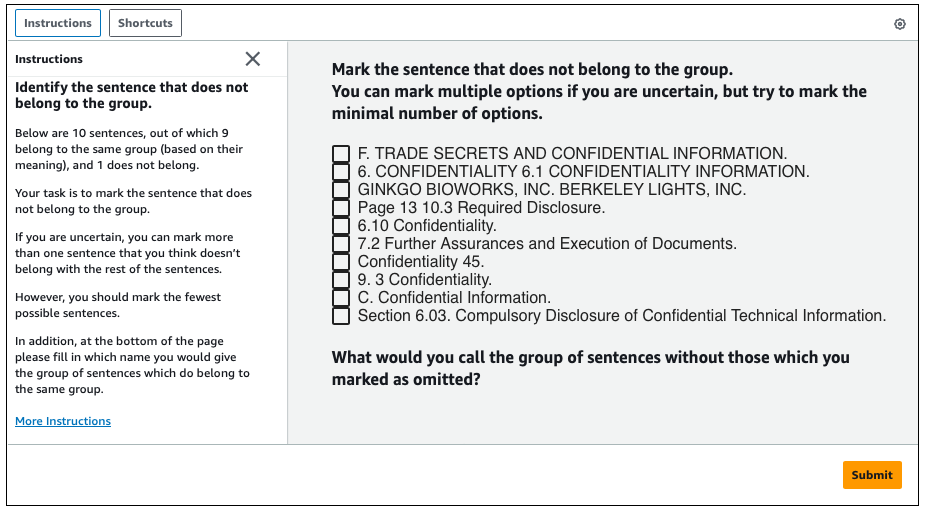}
    \caption{Screenshot of the human annotation interface for the header intrusion task. ``7.2 Further Assurances...'' and the rest are all headers describing a ``confidentiality'' section.}
    \label{fig:intruder-annotation}
\end{figure*}

Figure~\ref{fig:intruder-annotation} is a screenshot of the annotation interface for the header intrusion evaluation, As described in Section~\ref{sec:intruder-eval}. Annotators were chosen from English-speaking countries: USA, Canada, and UK. Mturk does not provide further demographic information about annotators.

\end{document}